\documentclass{article}
\usepackage{spconf,amsmath,epsfig}

\usepackage{multirow}

\usepackage[para,online,flushleft]{threeparttable}

\usepackage[noend]{algpseudocode}

\usepackage{algorithmicx,algorithm}
\algrenewcommand\algorithmicrequire{\textbf{Input:}}
\algrenewcommand\algorithmicensure{\textbf{Output:}}

\usepackage{fancyhdr}
\begin{document}\sloppy
\topmargin=0mm
\def\x{{\mathbf x}}
\def\L{{\cal L}}

\title{Spectral Analysis Network for Deep  Representation Learning and Image Clustering}
%
\name{Jinghua Wang$^1 $, Adrian Hilton$ ^2 $, Jianmin Jiang$^{1*} $}
\address{$^1$Research Institute for Future Media Computing, College of Computer Science and Software Engineering,\\  Shenzhen University, Shenzhen 518060\\
$^2$Centre for Vision, Speech and Signal Processing, University of Surrey, Guildford GU2 7XH, UK\\
$^*$Corresponding author: jianmin.jiang@szu.edu.cn}

\maketitle

\begin{abstract}
Deep representation learning is a crucial procedure in multimedia analysis and attracts increasing attention. 	
Most of the popular techniques rely on convolutional neural network and require a large amount of labeled data in the training procedure. However, it is time consuming or even impossible to obtain the label information in some tasks due to cost limitation.  Thus, it is necessary to develop unsupervised deep representation learning techniques.
This paper proposes a new network structure for unsupervised deep representation learning based on  spectral analysis, which is a popular technique with solid theory foundations. 
Compared with the existing spectral analysis methods, the proposed network structure has at least three advantages. Firstly, it can identify the local similarities among images in patch level and thus more robust against  occlusion.  Secondly, through multiple consecutive  spectral analysis procedures, the proposed network can learn more clustering-friendly representations and is capable to reveal the deep correlations among data samples. Thirdly, it can elegantly integrate different spectral analysis procedures, so that each spectral analysis procedure can have their individual strengths in dealing with different data sample distributions. Extensive experimental results show the effectiveness of the proposed methods on various image clustering tasks.  
\end{abstract}

\section{Introduction}

As one of the most fundamental techniques in multimedia analysis, clustering has been extensively studied \cite{icme-jia-convex}. The main goal of clustering is to categorize a  set of data samples into a number of clusters, so that the similar samples are in the same cluster and dissimilar ones in different clusters. 
Clustering techniques have a wide range of applications and achieve good performances\cite{icme-wang-co-reference}.

Spectral clustering \cite{Zelnik-Manor:2004:Self-tuning} is one of the most promising  techniques due to its elegance in theory and capability in exploring the intrinsic data structure \cite{Kang-arxiv-unified-2017}. It has been successfully  applied in various applications of multimedia analysis \cite{Liu2017Spectral}.

However, spectral clustering still has a number of  unsolved problems. Firstly, it is quite difficult to construct a proper affinity graph for a given dataset, though the affinity graph can influence the clustering results significantly. The main difficulties lie in the fact that it is not clear how to choose a proper similarity measurement and how to determine a suitable parameter for the chosen measurement. Secondly, there is no agreement in the choice of Laplacian matrix for eigenvector decomposition. Both of the two popular Laplacian matrices, i.e. symmetric normalized Laplacian matrix \cite{Ng01onspectral} and left normalized Laplacian matrix \cite{Shi2000-pami-Normalized},  have their own advantages and disadvantages.

In addition to clustering techniques, the representations of the data samples are of vital importance to achieve good clustering results.
Some research efforts target to learn proper  representations before performing the clustering procedure. 
The main goal is to improve the clustering performance by enhancing the intra-cluster similarity and reducing the inter-cluster similarity. 
Along with the popularity of deep learning techniques, an increasing number of researchers turn to convolutional neural network (CNN) to learn deep representations that are feasible for clustering \cite{Dizaji-iccv2017-deepClustering,Li2017DiscriminativelyBI,Shaham-iclr-2018SpectralNetSC,wang-accv}. 


Motivated by the significant success of deep learning, we extend the spectral analysis into multiple layers and propose a new spectral analysis network (SANet). SANet  learns deep representations based on multiple consecutive spectral analysis procedures and shows its strength in various image clustering tasks. 
Compared with the existing spectral analysis clustering, our SANet achieves the following three advantages.
Firstly, SANet provides a new method for deep representation learning and the learned representations are more suitable for clustering. 
Secondly, the proposd SANet can identify the local similarity between images by conducting multiple spectral analysis on image patches. In contrast, the existing spectral analysis methods only assess the similarity between image pairs holistically. This explains why SANet is more robust against occlusions.
Thirdly, the proposed SANet elegantly integrates multiple spectral analysis procedures in order to deal with data samples distributed differently. By employing different affinity graphs and Laplacian matrices,  different spectral analysis procedures can have their individual strengths in dealing with different data sample distributions.

%
%

\section{Related Work}
\label{sec-related-work}

The   spectral analysis clustering procedure is shown in Algorithm \ref{alg:spectral-clustering}. 
In comparison with other approaches, the spectral clustering method has at least three advantages. 
Firstly, the representations or embeddings (i.e. $ Q $ in Alg. \ref{alg:spectral-clustering}) in spectral clustering can be found analytically. 
Secondly,  spectral clustering analysis is able to handle non-convex datasets \cite{vonLuxburg2007-a-tutorial-on-spectral-clustering}.
Thirdly, spectral analysis has a solid theoretical foundation for further research. It can be derived from the view point  of graph cut, random walks, and matrix perturbation. However, there also exist a number of deficiencies in spectral analysis clustering.

\begin{algorithm}
	\caption{Spectral Clustering}
	\label{alg:spectral-clustering}
	\begin{algorithmic}[1]
		\Require 	$ n $ data points $ x_1, x_2, \cdots, x_n \in R^d $;  clustering number $ k $;
		\Ensure $ k $ clusters
		\State  Construct a  affinity matrix $ W \in R^{n \times n} $ between data points, with the degree matrix $ D \in R^{n\times n} $, where $ w_{ij} $ measures the similarity between $ x_i $ and $ x_j $;
		\State Compute the Laplacian matrix $ L=D-W $, where $ D_{ii}=\sum_{j=1}^{n} w_{ij}$;		
		\State Compute the  $ k $ eigenvectors $ Q_i (1 \leq i \leq k) $ of $ L $ associating with the $ k $ smallest eigenvalues, and denote them by $ Q=[q_i, q_2, \cdots, q_k] \in R^{n \times k}$;		
		\State For $ 1\leq i \leq n $, let $ y_i \in R^k $ be the $ i $th row of the matrix $ Q $, and apply $ k $-means to cluster the points $ y_i (1\leq i \leq n) $ to obtain the $ k $ clusters $ Cluster_j (1\leq j \leq k) $.		
	\end{algorithmic}
\end{algorithm}

Firstly, there is few theoretical analysis that leads us to a proper affinity matrix $ W $ for a given dataset. 
While  three different similarity measurements are  popularly used to construct the affinity matrix \cite{Zelnik-Manor:2004:Self-tuning}, i.e. $ k $-nearest-neighborhood, $ \epsilon $-nearest-neighborhood, and the fully connected graph, 
each of them can only deal with some but not all kinds of data sets.
While the $ k $-nearest-neighborhood strategy might break a connected component into several components, the $ \epsilon $-nearest-neighborhood strategy can not handle datasets with varied densities, and the fully connected affinity method  suffers from high  computational complexity.  In addition, it is quite difficulty to determine a proper parameter for these three strategies.

Secondly, the relevant research communities have not reached consensus on how to choose between different Laplacian matrices. 
The Laplacian matrix  has two popular extensions, i.e. symmetric normalized matrix $ L_{sym}=D^{-1/2}LD^{1/2}=I-D^{-1/2}WD^{1/2} $
and left normalized matrix $ L_{rw}=D^{-1}L=I-D^{-1}W $.
While Ng \cite{Ng01onspectral} adopted symmetric normalized Laplacian matrix and claimed superior performance, Shi \cite{Shi2000-pami-Normalized} and Luxburg \cite{vonLuxburg2007-a-tutorial-on-spectral-clustering} recommended left normalized matrix. Both normalizations have their individual advantages and disadvantages.

\section{Spectral analysis network}
\label{sec:the-proposed-method}

\subsection{Motivation}

Motivated by the success of convolutional neural network \cite{wang-eccv}, researchers also attempt to extend other techniques  to a network structure for deep representation learning \cite{chan-2015-tip-pcanet,Coates2012Learning-deep-kmeans,wang-csvt}. 
Following the similar spirit, we propose a new representation learning method via expansion of the concept in spectral analysis, and to the best of our knowledge, we are the first to explore the deep representation learning  based on the technique of spectral analysis.

It is widely recognized that the spectral features $ Y=\{y_i\}_{i=1}^n $ in Alg. \ref{alg:spectral-clustering} are more suitable for clustering than the original data points $ \{x_i\}_{i=1}^n $.   We consider to conduct   spectral analysis procedures on the spectral features $ Y $, in order to further improve the intra-cluster similarity and inter-cluster separability.

As previously mentioned, there are three different methods to construct the  affinity matrix, involving two different types of Laplacian matrices. As each of them has its own advantage, and it remains difficult to determine which one to use. To overcome this problem, we introduce a new deep  network structure that can integrate them together elegantly.

\begin{figure*}[htb]
	\centering	
	\includegraphics[width=0.7\linewidth]{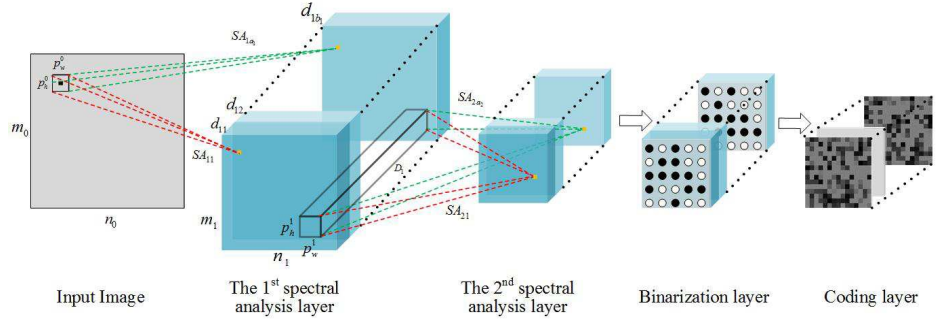}
	\caption{The structure of the proposed SANet. Besides input and output layers, there are four types of layers in the proposed network, spectral analysis layer, binarization layer, coding layer , and pooling layer (not shown in this figure).}
	\label{fig:network}
\end{figure*}

\subsection{Network Structure}

In this subsection, we propose a new network structure that can extract deep features or representations for the task of image clustering, as shown in Fig. \ref{fig:network}. The proposed network consists of four different types of layers, i.e. spectral analysis layer, binarization layer, coding layer, and pooling layer.

In the following, we show the details of these layers. 
For an image clustering task, we assume the image dataset  $I= \{ I_i | 1\leq i \leq N \}$ consists of $ N $  samples, and the image size is $ m_0\times n_0 $ with $ D_0 $ channels, i.e. $ I_i \in R^{m_0\times n_0\times D_0} $. 

(1) The first spectral analysis layer

The first spectral analysis layer extracts features from the image patches using the technique of spectral analysis. We call the output features of a spectral analysis layer as  spectral features.  In Alg. \ref{alg:spectral-clustering}, $ y_i $ denotes the spectral feature of sample $ x_i $.
Based on the theory of spectral analysis, we know that spectral features are more clustering-friendly than the original image patches.
In other words, the similarity is enhanced between a pair of neighboring patches and reduced between a pair of distant patches in the spectral feature space.

In the patch sampling procedure, we pad the image to include the border information. 
Around each of a subset (or all) pixels, we crop an image patch of size $ p_h^0 \times p_w^0 \times D_0 $, where $ D_0 $  represents the number of channels.
The mean patch is subtracted from each of the image patches to address the problem caused by illumination variations. For an image $ I_i $, we obtain a set of normalized image patches $ X_i^0=\{x_{ij}^0|1\leq j \leq n_p^0\} $. With  $ N $ images, the size of final image patch set is $ N\times n_p^0 $, and the patch set is $ X^0=\{X_1^0,X_2^0,\cdots, X_N^0\} $.

Given the image patch $ X_{ij}^0 $, the spectral analysis $ SA_{1t} (1 \leq t \leq b_1) $ extracts the first layer spectral feature $ F^1_{ij} $ with dimensionality of $ d_{1t}$. The parameter $ b_1 $ denotes the number of different spectral analysis procedures in the first layer. Note that, the  spectral analysis procedures can be different from each other in one or more of the following three aspects, including the affinity matrix, the Laplacian matrix, and  the optimization method. 
Let $ F_i^1 $ and $ F^1 $ denote the first layer spectral feature of $ X_i^0 $ and $ X^0 $, respectively. 
The spectral features of  one image (produced by $ SA_{1t}$) can be stacked into a $ m_1 \times n_1 \times d_{1t} $ matrix instead of being clustered directly.
With $ b_1 $ different kinds of spectral analysis procedures, we obtain the spectral features of an image with the dimensionality of $ m_1 \times n_1 \times D_1 $, where $ D_1=\sum_{t=1}^{b_1}d_{1t} $ sums the dimensionality of $ b_1 $ different set of spectral features.

(2) The second spectral analysis layer

The second spectral analysis layer  takes the concatenation of various spectral features produced by different  spectral analysis procedures (in the first layer) as the input. This layer further improves the discriminative ability of the  features.

Similar to the first layer, the second layer has two steps. Firstly, it samples feature patches on the output of the first layer, i.e. $ F^1 $. 
Secondly, let the feature patch set be $ X^1 $, this layer conducts spectral analysis on $ X^1 $ and produces the second layer spectral features $ F^2 $.

Let the size of the feature patches be $ p_h^1 \times p_w^1 \times D_1 $ as shown in the Fig. \ref{fig:network}. Each feature patch carries the information learned from a larger patch with size of $ (p_h^1+p_h^0-1) \times (p_w^1+p_w^0-1) \times D_0 $ in the original image. 
In addition, a feature patch also integrates the discriminative information learned by different spectral analysis procedures, which are suitable for the clustering of data samples with various distributions.

(3) The pooling layer

The pooling layer summarizes the neighboring spectral features in the same spectral map. 
We take  spectral features of the second layer  as an example. 
The spectral analysis $ SC_{2t} $ produces $ d_{2t} $ different feature maps of size $ m_2 \times n_2 $ for each image, and each feature map is associated with an eigenvalue.
The pooling operation is conducted inside each feature map. 
For a $ s_p \times s_p $ grid centered at a point, the pooling operation only keeps the strongest response in terms of absolute vale, which can be  mathematically expressed as
\begin{equation}
Pooling(G)=g_{kl} \qquad where \quad |g_{kl}|=\max \limits_{ij} |g_{ij}|
\end{equation}
where $ g_{ij} $ denotes the feature in the $ i $th row and $ j $th column of the feature grid $ G $. The pooling grids can be overlapped.

(4) The binarization layer

From the viewpoint of graph cut, the sign of spectral features (i.e. positive or negative) carries the cluster information \cite{vonLuxburg2007-a-tutorial-on-spectral-clustering}. 
For a two-cluster clustering task, we can simply take a single eigenvector in step $ 3 $ of Alg. \ref{alg:spectral-clustering}, and cluster  the data $ x_i $ into the first cluster if $ y_i >0 $ and  into the second cluster if $ y_i<0 $.
This observation explicitly shows the importance of the sign of the spectral features in the clustering task. Following the pooling layer, correspondingly, we use a binarization layer to binarize the spectral features, in which $ B_{ij} $ denotes the binary feature map corresponding to feature map $ F^2_{ij} $, and $ B_i $ denotes the feature maps of the $ i $th image.

(5) The coding layer

Following the binarization layer, a coding layer is added to transform the binary code into decimal numbers and thus make it feasible for the following clustering procedure. In this layer, we first partition the binary features of each image into different groups, and each group consists of $ L $ binary feature maps. We normally set $ L $ to be $ 8 $.
Let $ B_{ij}^k$ be the $ j $th $ (1\leq j \leq L) $ binary feature map  in the $ k $th group for the $ i $th image. 
At the position $ (u,v) $, we take the $ L $ binary features $ B_{ij}^k(u,v) $ and convert them into the  decimal using
$ C_i^k(u,v)=\sum \limits_{j=1}^L 2^{j-1}B_{ij}^k(u,v) $.
In this way, we obtain the $ k $th decimal feature map $ C_i^k $ for the $ i $th image. Note, the spectral features corresponding to smaller eigenvalues are assigned with larger weights, due to their stronger discriminant ability. With $ n_b $ binary feature maps, the coding layer produces  $ \lceil \left. n_b \middle/ L \right. \rceil
$ decimal feature maps. 
By setting $ L $ to be $ 8 $,  we obtain gray maps in the coding layer,  as shown in Fig. \ref{fig:network}. Finally, we can obtain the clustering results by conducting a simple K-means procedure on the output of the coding layer.

\section{Experiment}
\label{sec:experiment}

We conduct experiments  on handwritten digit image clustering, face image clustering, and fashion image clustering.
We adopt two popular standard metrics to evaluate the performance of different clustering methods, i.e. clustering accuracy (ACC) and nomarlized mutual information (NMI) \cite{cai-document-tkde}.

%
%
%
%
%

We compare our method (i.e. SANet) with a number of representative existing   clustering algorithms, including 
large-scale spectral clusteirng (SC-LS) \cite{Chen2011Large}, graph degree linkage-based agglomerative clustering (AC-GDL) \cite{Zhang2012Graph}, 
sepctral embedded clustering (SEC) \cite{Nie2011Spectral}, 
deep embedding clustering (DEC) \cite{Xie:icml2016:Unsupevisd-deep-learning},  
joint supervised learning (JULE) \cite{Yang-cvpr2016-joint}, 
and Deep embeded regularized clustering (DEPICT) \cite{Dizaji-iccv2017-deepClustering}.

\subsection{Handwritten digit image}

The USPS dataset\footnote{https://cs.nyu.edu/roweis/data.html} is a handwritten digits dataset consisting of $ 11,000 $ images. 
The MNIST dataset  \cite{mnist-lecun-1998-gradient} is one of the most popular image
datasets, consisting of $ 70,000 $ images.

For these two datasets, we use similar implementation details and take  MNIST as the example in the following. 
In the first layer, we sample $ 11 \times 11 $ image patches  with a stride of $ 5 $ both vertically and horizontally. With padding in the border, we sample $6 \times 6=36 $ image patches for an  $ 28 \times 28  $ image.


For $ k $-nearest-neighborhood affinity matrix construction, we set the parameter $ k $ to be $ 9 $ and $ 11 $. In the $  \epsilon$-nearest-neighborhood strategy, we have three different settings for the value of $ \epsilon $, i.e. $ 0.5\eta $, $ \eta $, and $ 2\eta $,  and $ \eta $ denotes the longest edge in the minimal spanning tree. 
We construct three different dense affinity matrices. One dense matrix is determined by the self-tunning method \cite{Zelnik-Manor:2004:Self-tuning}. The other two are constructed based on the Gaussian function $ w_{ij}=exp(-||x_i-x_j||^2/(2\sigma^2)) $ with the parameter $ \sigma $ equals to $ 0.1 $ and $ 0.01 $, respectively. Thus, we have $ 5 $ different sparse affinity matrices and $ 3 $ different dense affinity matrices.

A symmetric Laplacian matrix is computed from each of the affinity matrices to yield  spectral features. 
We apply Lanczos to obtain spectral features from sparse affinity matrices, and mini-batch anlaysis to derive spectral features from dense affinity matrices. For each of the Laplacian matrix, we calculate $ 64 $ dimension of spectral features. 
To summary, the spectral features produced by the first layer is of size $ 6 \times 6 \times 512 $ for each image, with $ 64 $ dimensional spectral features for each $ 8 $ different Laplacian matrices.

\begin{figure}[htb]
	\centering
	\includegraphics[width=0.65\linewidth]{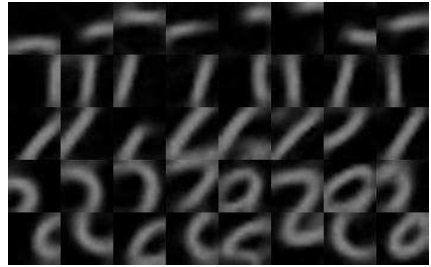}
	\caption{The typical visual patterns in the MNIST dataset}
	\label{fig:mnist-patches}
\end{figure}

The first layer can learn the typical visual patterns in the image dataset. To show this, we conduct  a K-means procedure based on the first layer spectral features and visualize the cluster centers in Fig. \ref{fig:mnist-patches}. As  seen, while the first three rows represent  lines in different angles and positions, the last two rows represent  different curve shapes appear in the digit images.  
A proper combination of these visual patterns can produce a digit image.

In the second layer, we sample feature patches with the size of $ 4\times 4 \times 512 $.  With a stride of $ 1 $ and no feature padding, we obtain  $ 3\times3=9 $ feature patches for each of the $ 6 \times 6 \times 512 $ first layer spectral features. In other words, each image is associated with $ 9 $ feature patches with dimensionality of  $ 4\times 4 \times 512 $.
This layer also uses both sparse affinity matrices and dense affinity matrices in the spectral analysis procedures.
With a symmetric Laplacian matrix employed, therefore, this layer has $ 4 $ different spectral analysis procedures altogether.
As each spectral analysis produces $ 16 $ dimensional features, the second spectral analysis layer produces $ 64 $ feature maps for each image and each feature map is of the size $ 3\times3 $. 
Following the binarization and coding layer (with $ L=8 $), we now have $ 64/8=8 $ coding feature maps, each of which has the size of $ 3 \times 3 $. In other words, the dimensionality of the  features  for the final $ k $-means procedure is $ 72 $.

Tab. \ref{tab:handwritten-clustering-performance} lists the clustering performances of the proposed SANet, in comparison with the benchmarks. In terms of both ACC and NMI, 
our proposed SANet achieves the best performances, which indicates that the proposed network can learn feasible deep features for the clustering task.

\begin{table}[htb]
	\scriptsize 
	\centering
	\caption{The clustering performances of different methods on handwritten digit image datasets}
	\begin{tabular}{cllllllll} 
		\hline
		\multicolumn{2}{l}{{\tiny Methods}}       &  {\tiny SC-LS}  & {\tiny AC-GDL} &  {\tiny SEC}   &   {\tiny DEC}& {\tiny JULE}& {\tiny DEPICT}& {\tiny SANet}    \\ 
		\hline
		\multirow{2}{*}{mnist} & acc & 0.311 &  0.113  & 0.804 & 0.844& 0.959& 0.965& 0.970  \\ 
		& nmi &  0.416  & 0.017    & 0.779 &   0.816& 0.906& 0.917& 0.923  \\ 
		\hline
		\multirow{2}{*}{usps}  & acc &  0.308  & 0.867    & 0.544 &    0.619& 0.922& 0.964& 0.976  \\ 
		& nmi   & 0.726 & 0.824    & 0.511 & 0.586& 0.858& 0.927& 0.936  \\ 
		\hline
	\end{tabular}
	\label{tab:handwritten-clustering-performance}
\end{table}

\subsection{Occluded Face Image Clustering}

\begin{figure}[htb]
	\centering
	\includegraphics[width=0.65\linewidth]{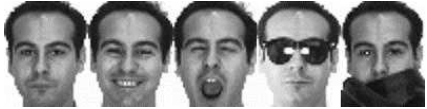}
	\caption{The example images from the AR dataset}
	\label{fig:ar-images}
\end{figure}

The AR dataset \cite{ar-dataset} consists of more than $ 4,000 $ frontal face images from $ 126 $ people. 
The face images were captured under different conditions introduced by facial expression, illumination variation, as well as disguise (sunglass and scarf). The images were captured in two sessions (with an interval of two weeks). 
For the convenience of implementation without generality, we crop the image to be the dimension of $ 50 \times 40 $. 
Fig. \ref{fig:ar-images} shows five example images from the same person.

Following the same design as for the clustering of digit images, we apply the proposed network structure shown in Fig. \ref{fig:network} for face image clustering. However, some of the implementation details need to be different.  Firstly, for face image clustering, we sample patches of size $ 15 \times 15 $ with stride of $ 7 $. The patches are  larger than the ones used in the handwritten image dataset, in order to allow the typical visual patterns to cover meaningful parts of the faces. Secondly, we use the Nystrom approximation  (a different method from the digit image clustering experiment) to compute the spectral features from the dense affinity matrices,  in which 20\% of all the patches are randomly selected as the seen data. 

Tab. \ref{tab:face-image-clustering-performance} summarizes the experimental results, from which it can be seen that our proposed SANet outperforms all the $ 6 $ benchmarks selected out of the existing clustering algorithms. Further examinations of the experimental results also reveal that,  compared with all other spectral analysis-based clustering methods, the proposed method achieves the advantage  in dealing with the occluded face images, due to the fact  that the network allows us to identify the local similarity between the face images in patch-level. 

\begin{table}[htb]
	\scriptsize 
	\centering
	\caption{The clustering performances of different methods on AR face image dataset}
	\begin{tabular}{cllllllllllllll} 
		\hline
		Methods   & SC-LS & AC-GDL & SEC   &   DEC& JULE& DEPICT& SANet    \\ 
		\hline
		ACC &   0.276 & 0.356  & 0.356 &  0.459& 0.561& 0.504& 0.637  \\ 
		NMI &   0.338 & 0.421   & 0.319 &  0.437& 0.501& 0.473& 0.586  \\ 
		\hline
	\end{tabular}
	\label{tab:face-image-clustering-performance}
\end{table}

\subsection{Fashion Clustering}

We carry out another phase of experiments to
cluster the fashion images into different styles on the HipsterWars \cite{hipsterwars-kiapour-eccv2014}. 
This dataset consists of $ 1,893 $ fashion images, each associating with one of five style categories, i.e. hipster, bohemian, pinup, preppy, and goth. The numbers of images in these five categories are $ 376 $, $ 462 $, $ 191 $, $ 437 $, and $ 427 $ respectively. 

The first spectral analysis layer samples image patches of size $ 32 \times 32 $. We use $ 8 $ different spectral analysis procedures as in the digit image clustering to learn the spectral features. 
The main goal of the first layer is to  discover the typical visual patterns that appear  in many fashion images. In the second layer, we only use sparse affinity matrix constructed by the $ k $-nearest-neighborhood, with the parameter $ k $ equals to $ 5, 7, 9 $ and $ 11 $. We adopt both the symmetric normalized matrix and the left normalized matrix, and apply the Lanczos method for Laplacian matrix decomposition to produce spectral features.

\begin{table}[hbt]
	\footnotesize
	\centering
	\caption{The clustering performances of different methods on HipsterWars dataset}
	\label{tab:hipsterwards-nmi-acc}
	\begin{tabular}{lllll}
		\hline
		& StyleNet & ResNet & PolyLDA & SANet \\
		\hline
		ACC & 0.39     & 0.30   & 0.50    & 0.54  \\
		NMI & 0.20     & 0.16   & 0.21    & 0.20 \\
		\hline
	\end{tabular}
\end{table}

\begin{figure}[htb]
	\centering	\includegraphics[width=0.65\linewidth]{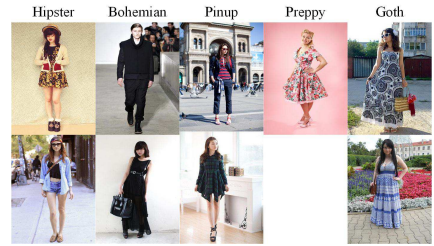}
	\caption{The fashion images nearest to the cluster centers of the five different styles}
	\label{fig:representativeFashion}
\end{figure}

Tab. \ref{tab:hipsterwards-nmi-acc} summarizes the clustering results of the proposed method and three existing state of the art benchmarks, including StyleNet \cite{Simo-fashion-cvpr2016}, ResNet \cite{He2016Deep}, and PolyLDA \cite{Hsiao-iccv2017}. As seen, our proposed method performs better than
existing benchmarks in terms of accuracy. 
For the convenience of further examination and analysis, 
Fig. \ref{fig:representativeFashion} illustrates some image samples, which are nearest to the cluster centers.

\section{Conclusion}
\label{sec:conclusion}

In this paper, we proposed a new network structure, i.e. SANet,  based on the technique of spectral analysis. This provide one more method 
for deep representation learning, in addition to the popular convolutional neural network.
The newly proposed network structure has four type of layers.

\section*{Acknowledgment}

The authors wish to acknowledge the financial support from: (i) Natural Science Foundation China (NSFC) under the Grant No. 61620106008; (ii) Natural Science Foundation China (NSFC) under the Grant No. 61802266; and (iii) Shenzhen Commission for Scientific Research \& Innovations under the Grant
No. JCYJ20160226191842793.

\bibliographystyle{IEEEbib}
\bibliography{1222-bib}

\end{document}